\documentclass[preprint,12pt]{elsarticle}

\usepackage{amssymb}
\usepackage{amsmath}

\usepackage{amsfonts}
\usepackage{algorithmic}
\usepackage{longtable}
\usepackage{array}
\usepackage{lipsum} 
\usepackage{booktabs} 
\usepackage{tabularx} 
\usepackage{multirow}   
\usepackage{caption}
\usepackage{textcomp}
\usepackage{stfloats}
\usepackage{cite}
\usepackage[normalem]{ulem}
\usepackage{url}
\usepackage{verbatim}
\usepackage{graphicx}
\usepackage{colortbl}
\usepackage{makecell}
\usepackage[colorlinks=true, linkcolor=red, citecolor=blue, urlcolor=magenta]{hyperref}
\usepackage{booktabs}   
\usepackage{tabularx}   
\usepackage{ragged2e}   
\usepackage{geometry}   
\usepackage{xcolor}     
\usepackage{ulem}       
\usepackage{graphicx} 
 \usepackage{xcolor}    
\usepackage{array}     

\usepackage{pgfplots}
\usepackage{tikz}
\usepackage{CJKutf8}
\def\BibTeX{{\rm B\kern-.05em{\sc i\kern-.025em b}\kern-.08em
    T\kern-.1667em\lower.7ex\hbox{E}\kern-.125emX}}
\usepackage{balance}

\journal{Expert Systems with Applications}

\begin{document}
\begin{frontmatter}

\title{ Low-light Image Enhancement via Multi-scale Attention combined with Fourier Transform}
\author[ecust]{Wenbin Du}
\author[ecust]{Jian Long\corref{cor1}} 
\ead{longjian@ecust.edu.cn}
\author[tju,t]{Zhu Cao\corref{cor1}}  
\cortext[cor1]{Corresponding authors} 
\address[ecust]{Key Laboratory of Smart Manufacturing in Energy Chemical Process, Ministry of Education, East China University of Science and Technology, Shanghai, 200237, China}
\address[tju]{College of Electronics and Information Engineering, Tongji University, Shanghai 201804, China}
\address[t]{Shanghai Research Institute for Intelligent Autonomous Systems, Tongji University, Shanghai 201210, China}
\ead{caozhu@tongji.edu.cn}

\begin{abstract}
Low-light image enhancement (LLIE) aims to improve image quality and clarity in diverse and demanding low-illumination environments. However, existing deep learning-based LLIE methods struggle to accurately capture real-world illumination and restore texture details, largely because their algorithmic strengths remain underutilized. To address these issues, we present a supervised frequency domain deep learning network for LLIE, named multi-scale attention combined with the Fourier transform (MSFT) which adopts a U-shaped, one-stage architecture that infuses guidance from low-light images into the network by channeling it through multi-scale attention. We further fuse the amplitude information from priori channels with that of the low-light image in MSFT's self-created module, and carry out multi-scale guidance along with the network. Subsequently, to better enhance the faint feature, such as fine content and textures, and to better fuse global context confidence in the decoding stage, we separately introduce a multi-shape synergistic attention and a lightweight network that effectively integrate information in high-dimensional space to embed into the superlative feature space channel containing rich texture information. Extensive experiments conducted on LOL, SID, SMID, and SDSD datasets demonstrate that MSFT significantly outperforms state-of-the-art competitors. For example, compared with Retinexformer, our method achieves a peak signal-to-noise ratio of up to 41.76 decibels on the SDSD-outdoor dataset with an increase of 11.92 decibels and a structural similarity index of 0.988 with a 13.80\% improvement. 
\end{abstract}
\begin{keyword}
Low-light image enhancement, Fourier transform, Multi-scale attention, Amplitude information, Multi-shape synergistic attention.
\end{keyword}
\end{frontmatter}

\section{Introduction}
Low-light Image Enhancement (LLIE) is dedicated to obtaining high-quality normal-light images for observation and downstream tasks, e.g., night image semantic segmentation \citep{TAN2025117265, s22145312, 8569387}, dark object detection \citep{yin2023peyolopyramidenhancementnetwork, 10.1007/978-3-030-58589-1_21}, dark face detection \citep{wang2021hlafacejointhighlowadaptation, 10.1609/aaai.v38i6.28339} and nighttime object hallucination \citep{Zhang2024Advancing, 2020Comparative}. In the real world, the acquisition of low-light images is inherently associated with challenging environmental conditions such as insufficient nighttime indoor illumination and dramatic variation of the brightness of the outdoor scene \citep{rea2016scene}, leading to the occurrence of luminance degradation \citep{s21093182} and substantial image blur \citep{zhang2018learning}. 
Visual enhancements in low-light images require not only improvements in image brightness and contrast \citep{patel2019review}, but also effective texturing techniques for detailed enhancement in low-light background regions. However, conventional LLIE techniques, e.g. histogram equalization methods \citep{lee2013contrast, liu2021benchmarking}, Retinex-based optimization methods \citep{parthasarathy2012automated, wu2022uretinex}, and various hand-made traditional methods \citep{fu2016weighted, 1284395}, which are heavily dependent on manual adjustments, are laborious and time consuming. Generally neglecting the processing of missing information in image black spots, these methods fail to meet the demands for human visibility and may even compromise the performance of the model.

Fortunately, the rapid development of deep learning has promoted the implementation of numerous deep learning-based models for LLIE. These models mainly use deep neural networks \citep{SCHMIDHUBER201585} frameworks to learn to manipulate brightness, color, tone, and contrast information from normal light maps to improve image quality. However, mainstream deep learning methods \citep{zhang2021rellie, 9010274, DBLP:journals/corr/LoreAS15} are limited in their ability to adaptively adjust the brightness distribution based on the texture information present in an image, as they fail to take into account the regional distribution of the texture information. Some approaches integrate semantic information across various image regions. For instance, Wu et al. \citep{wu2023learning} proposed a semantic-aware knowledge-guided framework to assist LLIE models in learning semantic priors from semantic segmentation models. However, these methods generally require substantial computational resources. Various dynamic enhancement methods exist to improve the perceptual quality of image pixels while avoiding reliance on semantic priors. Among these, the Signal-to-Noise-Ratio (SNR)-aware technique has garnered significant attention for its effectiveness \citep{xu2022snr}, which effectively enhances low SNR areas within an image and achieves a spatially variable enhancement. Although SNR-aware optimization demonstrates superior performance in achieving high signal-to-noise ratios, it fails to capture deep-scale information and inadequately enhances background regions containing auroral foreground elements.
\begin{figure}[!t]
\centering
\includegraphics[width=1.0\textwidth]{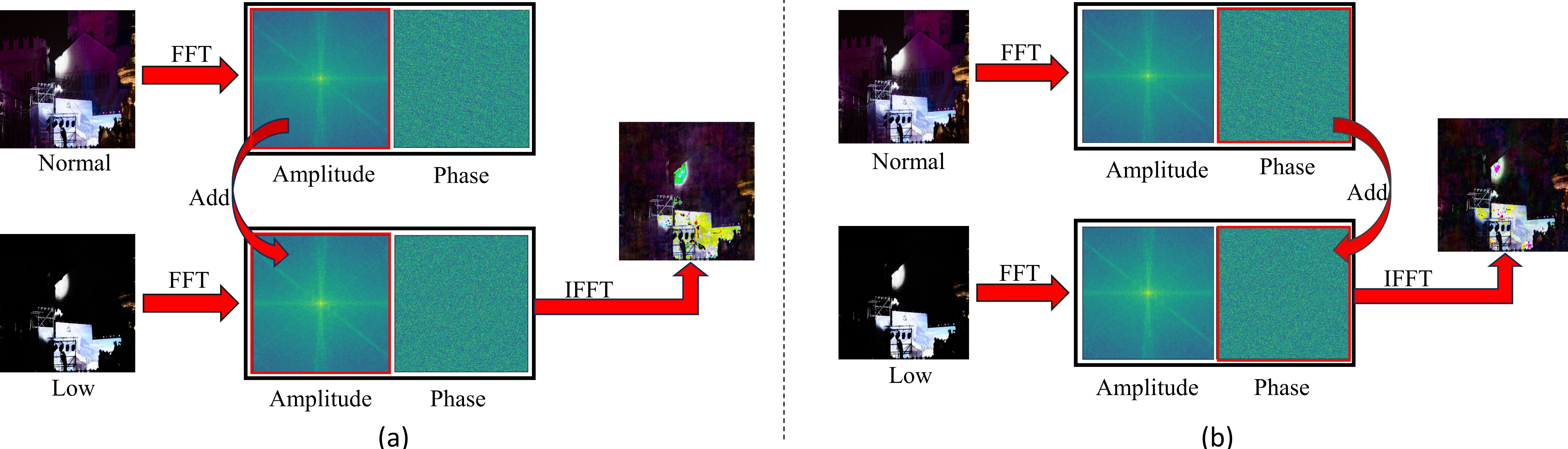}
\caption{\protect (a) Fourier transforms of of amplitude addition. (b) Fourier transforms of phase addition.}
\label{fig2}
\end{figure}

To address these issues, diffusion models exhibit exceptional capabilities in generating highly realistic and intricately detailed images. Most diffusion-based networks \citep{hou2024global,jiang2023low} treat low-light images as image generation or image restoration tasks, which adopt supervised learning approaches with large-scale paired data to restore detailed image textures and optimize contrast. Consequently, pre-networks are commonly embedded in diffusion models to obtain prior images that need to be repaired beforehand \citep{10890851, 10678573}. For example, Jiang et al. \citep{jiang2025lightendiffusion} proposed an unsupervised generation framework, LightenDiffusion, which takes the reflectance map of the low-light image and the illumination map of the normal-light image as input to the diffusion model for unsupervised restoration. Zhou et al. \citep{zhou2024low} first proposed a pipeline to guide VLMs in evaluating various visual attributes of low-light images, quantifying the assessment to generate global and local perceptual priors. However, these models are highly dependent on prior knowledge acquired by pre-trained networks, and the recovery process based on diffusion models usually relies on an assumed degradation process. These constraints pose challenges to the practical implementation of diffusion-based methods in real-world low-light environments, where the exact nature of degradations is often ambiguous and cannot be accurately modeled. This limits their broader applicability in real-world scenarios.

To highlight the texture details of the low-light background, while ensuring precise illumination recovery, we propose a Multi-Scale attention combined with the Fourier Transform (MSFT) method. This approach leverages the observation that the frequency domain amplitude encodes the main brightness information, while the phase components relate to noise and structural details, as shown in Figure \ref{fig2}. Figure \ref{fig2}(a) shows the amplitude spectrum of a normal light image decomposed by FFT, which is added pixel-by-pixel to the amplitude spectrum of a low-light image. Figure \ref{fig2}(b) presents the phase superposition of the normal light image and the low-light image using the same decomposition and addition methods as in Figure \ref{fig2}(a). Both images are then reconstructed via inverse Fourier transform. The visualization results show that the former successfully achieves a brightness enhancement effect similar to that under normal lighting conditions, which provides a solid, realistic basis for the conception that we can prevent overexposure and enhance dim backgrounds at the same time. To make the algorithm a transferable independent system, its guiding information is constructed by concatenating the maximum gray value prior map with the low-light image channel, thereby forming a four-channel feature brightness map. Based on this, the enhancement weights can be adaptively adjusted according to the maximum gray value in the low-light area, thereby preventing the occurrence of overexposure and underexposure phenomena. Specifically, we design a multi-head self-attention module guided by the Fourier frequency-guided transform (FFG-MSA), which adds the amplitude of the four-channel feature brightness mape to the amplitude of the low-light image in the spatial domain as guidance information. In addition, since the amplitude patterns of images at different resolutions are similar, the guidance information can be performed at different scales. In detail, this guidance information is initially directed across multiple channel scales, followed by utilizing plug-and-play Multi-Shape Synergistic Attention (MSSA) (comprised of Spatial and Channel Synergistic Attention (SCSA) \citep{si2024scsa} mechanism and Multi-Shape Attention (MSA) \citep{10713101}) to extract deep-scale texture information at the highest channel scale, and ultimately integrating the extracted texture-level spatial information for fusion. The combined module of MSSA for extracting guiding information in high-dimensional space and the remaining extraction modules have the greatest effectiveness because it can not only fully integrate the sparse information within multiple channels and homogenize the channel information density, but also effectively guide the re-calibration of different semantic information and alleviate the differences in multiple semantic information. Meanwhile, inspired by the framework of IA-LLIE \citep{pan2024illumination}, we align the information of feature brightness maps at different scales with the decoded part information at the corresponding scales to maintain the consistency of the enhanced content. Our contributions are summarized as follows.

\begin{itemize}
\item We propose a method, MSFT, a two-stage integrated network combining the Fourier transform with a CNN-Transformer Cross-mixed U-net framework. This hybrid framework combines CNN efficiency in capturing local features with the Transformer's ability to model the global context, achieving a more powerful and efficient feature representation. Additionally, the symmetrical encoding-decoding structure is conducive to achieving a balance between a precise understanding of the global context of the image and the perfect reconstruction of local details, thereby enabling the efficient generation of high-fidelity and richly detailed high-quality results. Furthermore, in order to fully integrate the discrete and sparse features in the high-dimensional space, we introduced the MSSA module in the highest-dimensional one-way guidance space. 
\item We design a self-attention mechanism, FTG-MSA, which serves as the core boost recovery module embedded into multiple scales of MSFT. By incorporating frequency-domain information, the cross-attention mechanism can dynamically and selectively draw the most relevant features from the source data during enhancement, thereby enriching the image content. We conducted extensive experiments on five representative datasets: LOL, SID, SMID, and SDSD, and the subsets of the LOL and SDSD datasets were further divided. The result demonstrates that our method MSFT is superior to the existing state-of-the-art (SOTA) methods and exhibits strong generalization capability. 
\end{itemize}

\section{Related work}
\subsection{Low-light Image Enhancement.}
\textit{Traditional methods.} LLIE aims to improve the quality of low-visibility images that suffer from poor lighting conditions. Traditional methods for low-light enhancement mainly rely on gamma correction 
 \citep{Veluchamy_Subramani_2019, rahman2016adaptive}, histogram equalization \citep{abdullah2007dynamic, Cheng_Shi_2004, pizer1987adaptive}, and traditional Retinex theory \citep{Lee2013Ada, Wang2008Ima} to enhance low-light images. Jobson et al. \citep{jobson1997multiscale} improved the single-scale Retinex to multi-scale Retinex (MSR). By combining multiple-scale Gaussian filtering kernels, MSR can better handle the problem of uneven illumination in images while achieving dynamic range compression, color consistency, and brightness reproduction. Ma et al. \citep{ma2017multi} proposed Multi-Scale retinex with Color restoration by adding a color restoration factor based on MSR to adjust the color distortion caused by the enhancement of local contrast in the image, thus highlighting the information in relatively dark areas and eliminating the color distortion defect in the image. However, these methods largely overlook the influence of lighting conditions and lack effective strategies to address noise. In recent years, deep learning LLIE methods have commonly achieved better results than traditional hand-crafted methods. 
 
 \textit{The methods based on the iterative learning.} The deep learning approach designs networks based on specific learnable parameter matrices and optical theories such as Retinex. This enables the network to learn ideal image enhancement priors, thereby accurately achieving low-light enhancement, exposure stabilization, and other functions. Wei et al. \citep{wei2018deep} propose a deep Retinex-Net that integrates a decomposition network based on Retinex theory with the deep enhancement network. Ma et al. \citep{ma2022toward} established a cascaded illumination learning process with weight sharing to improve exposure stability and reduce computation cost.Wu et al. \citep{wu2022uretinex} proposed a Retinex-based depth-expansion network (URetinex-Net), which transforms the optimization problem into a learnable network, decomposing the low-light image into a reflectance layer and a lighting layer. 
 
 \textit{The methods based on deep neural networks.} Utilizing the decomposition problem as an implicit prior regularization model, with the continuous advancement of deep learning research, a series of high-performance fixed backbone networks, such as Generative Adversarial Networks, Transformers, and Diffusion models, have emerged and are increasingly being applied in the field of low-light image enhancement. These network-based approaches, which perform pixel-level processing, demonstrate significantly enhanced application potential. Jiang et al. \citep{jiang2021enlightengan} designed an unsupervised generative adversarial network to enhance images with unpaired training. PyDiff\citep{zhou2023pyramiddiffusionmodelslowlight} introduces a pyramid diffusion method coupled with a global corrector to tackle the slow inference speed and global degradation issues of traditional diffusion models in LLIE tasks, thereby significantly enhancing both the quality and efficiency of image enhancement. LightenDiffusion\citep{jiang2024lightendiffusion} incorporates the Retinex decomposition mechanism within the latent space to reconstruct the reflectance map. Chan et al. \citep{chan2024AnlightDiff} proposed Anlightendiff, which stabilizes and enhances the process by dynamically adjusting the diffusion anchoring mechanism and the sampler. In addition, they designed a feature-aware loss specifically for diffusion models to comprehensively utilize different loss functions in the image domain. Although these methods demonstrate strong generalization performance and are capable of handling most complex scenes, they struggle to accurately address the intricacies associated with image noise and areas affected by insufficient exposure conditions. The Vision Transformer (ViT) \citep{alexey2020image} and its variants have gained significant popularity in the field of computer vision, which is attributed to the substantial advantages of the global attention mechanism over the convolutional layer in capturing long-range dependencies and its exceptional performance. For example, SNR-Net \citep{xu2022snr} employs SNR-aware transformers that tailor its operations based on regional SNR, performing long-range modeling on regions with extremely low SNR while reserving short-range operations for others. However, methods employing a fixed ViT structure, such as SNR-Net, are constrained by their globally uniform enhancement mechanism. They struggle to adaptively handle low-light images with uneven illumination, thus easily leading to local over-exposure. Cui et al. \citep{pan2024illumination} propose a lightweight and fast illumination adaptive Transformer, which utilizes ViT to restore the normally lit sRGB image locally and globally. Retinexformer \citep{cai2023retinexformer} used the illumination-guided transformer to restore the illumination information corrupted by a Retinex-based framework. Pan et al. \citep{FAN2025104276} designed an illumination-aware transformer network that incorporates different scales of illumination features to address the problem of uneven illumination distribution in low-light images. Dong et al. \citep{dong2025sallie} proposed SG-LLIE to address the instability of the network's inherent structure and the difficulty in retrieving robust semantics from severely damaged images, and introduced structural priors to adjust the later enhancement process. The aforementioned application of a ViT in low-light image enhancement demonstrates its significant potential for low-level visual tasks.
 
\subsection{Fourier Transform In Frequency Domain.}
Recently, the frequency domain-based Fourier transform has been gradually applied in computer vision tasks. In medical image segmentation \citep{ewaidat2024frequency} and image enhancement \citep{li2013image, wang2023fourllie}, by properly decomposing the frequency domain, high-definition images can efficiently perform more complex calculations. Huang et al. \citep{huang2022deep} identify that the amplitude in Fourier space encodes brightness, while the phase captures structural details. This insight leads us to propose the deep Fourier-based Exposure Correction Network (FECNet), which employs dedicated amplitude and phase sub-networks to progressively reconstruct the lightness and structure of an image. UHDformer \citep{li2023embedding}, incorporating the Fourier transform into a cascading network, effectively enhances the image and avoids the amplification of noise. Furthermore, channel-dimension Fourier transform learning \citep{Li2023ICLR}, applying the Fourier transform on the channel dimension, provides an effective space to adjust the global representation. The aforementioned application of a ViT in low-light image enhancement demonstrates its significant potential for low-level visual tasks. FourierDiff \citep{lv2024fourier} is the first to incorporate Fourier priors into a pre-trained diffusion model to effectively address the simultaneous degradation of luminance and structural elements by zero-shot learning. Zhang et al. \citep{zhang2024dmfourllie} proposed a Dual-Stage Multi-Branch Fourier LLIE framework to address color distortion and noise in LLIE that uses the Fourier transformation, which involves amplifying amplitude components and copying phase components. As shown in Table \ref{tab:low_light_enhancement_methods} of the above-mentioned related work.  Inspired by the above work in the frequency domain, we extract multi-scale frequency features to insert into the corresponding scale self-attention.

\hyphenation{
    ar-chI-tec-ture en-hance-ment de-com-po-si-tion re-con-struc-tion
    il-lu-mi-na-tion pre-dic-tion pa-ra-me-ters gen-er-al-i-za-bil-i-ty
}
\onecolumn
{
    \small	
    \fontsize{8}{9}\selectfont
    \renewcommand\arraystretch{1.1}	
        \begin{longtable}{>{\color{black}}p{2.5cm}>{\color{black}}p{4cm}
				>{\color{black}}p{3cm}>{\color{black}}p{3cm}>{\color{black}}p{3cm}
				}
        \caption{Comparison of Related Low-Light Enhancement Methods}\\  
        \label{tab:low_light_enhancement_methods}\\
        \toprule
        Method        & Algorithmic model                                                                 & Datasets                                                                 & Advantages                                                              & Limitations
        \\ 
        \midrule
        \addlinespace
        \endhead
        {RetinexNet \citep{wei2018deep}}     & {Decomposition → Adjustment → Reconstruction}                                        & {LOL and RAISE}                                                                  & {Integrates Retinex theory with neural networks}                            & {Color distortion in color images}
        \\
        \addlinespace
        {Self-Calibrated Illumination (SCI) \citep{ma2022toward}}          & {Illumination Learning with Weight Sharing + Self-Calibrated Module + Noise Removal Modules + Unsupervised Training Loss}                                     & {MIT and LSRW}                                                       & {Stable exposure, fast, flexible, and robust}  & {Low generalization ability}                                                                        \\ 
        \addlinespace
        {URetinex-Net \citep{wu2022uretinex}}           & {Design learnable modules for initialization based on data dependencies, expansion optimization, and specification of lighting enhancement}                                                      & {LOL and SICE}                                 & {Realize the noise suppression and detail preservation of the final decomposition result.}                    & {Lack of effective processing for areas with uneven lighting}
        \\ 
        \addlinespace  
        EnlightenGAN \citep{jiang2021enlightengan}  &Global-Local Discriminators + U-Net Generator                                       & MEF, NPE, LIME, VV, DICM, BBD-100k, and LOL                                    & Global-local synergy; balances authenticity and accurac                  & Generalizability limited by training dataset  \\
        \addlinespace
        PyDiff \citep{zhou2023pyramiddiffusionmodelslowlight}      & Pyramid structure + Diffusion model                                          & LOL, LSRW
        , DICM, NPE, and VV                                              & Adaptability in multiple scenarios, rich content restoration                                    & High consumption of computing resources                            \\ 
        \addlinespace
        LightenDiff \citep{jiang2024lightendiffusion}       & Retinex decomposition of the latent space + illumination restoration of the unsupervised diffusion model                                          & LOLv2-real, LOLv2-synthetic                                              & Prevent the occurrence of global degradation and improve computational efficiency                                    &Poor adaptability of dynamic scenes                            \\ 
        \addlinespace
        Anlightendiff \citep{chan2024AnlightDiff}      & A dynamical regulated diffusion anchoring mechanism                                         &LOL, VE-LOL, LOLv2-real, NPE, LIME, and VV                                            & Dynamically adjust the diffusion process to adapt to complex scenarios and suppress various noises                                    &The distortion of color content and the insufficient ability to restore details in high-contrast scenes                           \\ 
        \addlinespace
        SNR-Net \citep{xu2022snr}      & Transformer + CNN                                                                   & LOL series, SID, SMID, and SDSD (indoor/outdoor)                               & Spatially adaptive enhancement                                          & Insufficient recovery of near-black regions               \\ 
        \addlinespace
         IA-LLIE \citep{zhang2021rellie}      & Transformer + CNNs + U-net architecture     & LOLv1, LOLv2-real, LOLv2-synthetic,  LSRW                              & Superior lighting adaptation, true-to-life color recovery & Insufficient noise robustness       \\ 
        \addlinespace
        Retinexformer \citep{cai2023retinexformer} & One-stage Retinex framework + illumination-guided CNNs-Transformer hybrid architecture                 & LOL series, SID, SMID, and SDSD (indoor/outdoor)              & Avoids overexposure; preserves details and colors         & Reversal distortion in bright areas \\     
        \addlinespace
        UHDformer \citep{li2023embedding}    & Cascade network + Fourier Transform                     &  LOLv1, LOLv2, and UHD-LL                                     & Excellent cross-scene generalization, superior denoising capability, and refined enhancement tailored for high-resolution images                    & Not suitable for dynamic scenes \\ 
        \addlinespace
        SG-LLIE \citep{dong2025sallie}      & Structure-Guided Transformer block + Hybrid Structure-Guided Feature Extractor + U-net architecture                  & LOL-v1, LOLv2-real, and NTIRE 2025 LLIE                   & Robustness of image texture structure                & Poor noise robustness\\
        \addlinespace
        FourierDiff \citep{lv2024fourier}         & Fourier priors-guided diffusion model +  Spatial-frequency alternating optimization & LOL-blur, RealBlur & Clear, consistent, and independence of paired images.      &  Fails without content priors and under extremely low illumination                                    \\ 
        \addlinespace
        DMFourLLIE \citep{zhang2024dmfourllie}         & Fourier reconstruction + Fourier convolution + multi-scales convolution        & LSRW-Huawei, LOL-v2-real, LOL-v2-synthesis, LIME, VV, DICM, NPE, and MEF                                                         & High-fidelity fine-grained texture and spatial structure           & Only suitable for reconstruction in static scenes                                                     \\ 

        \bottomrule
    \end{longtable}
}

\section{Method}
\begin{figure}[htbp]
\centering
\includegraphics[width=1.0\textwidth, keepaspectratio]{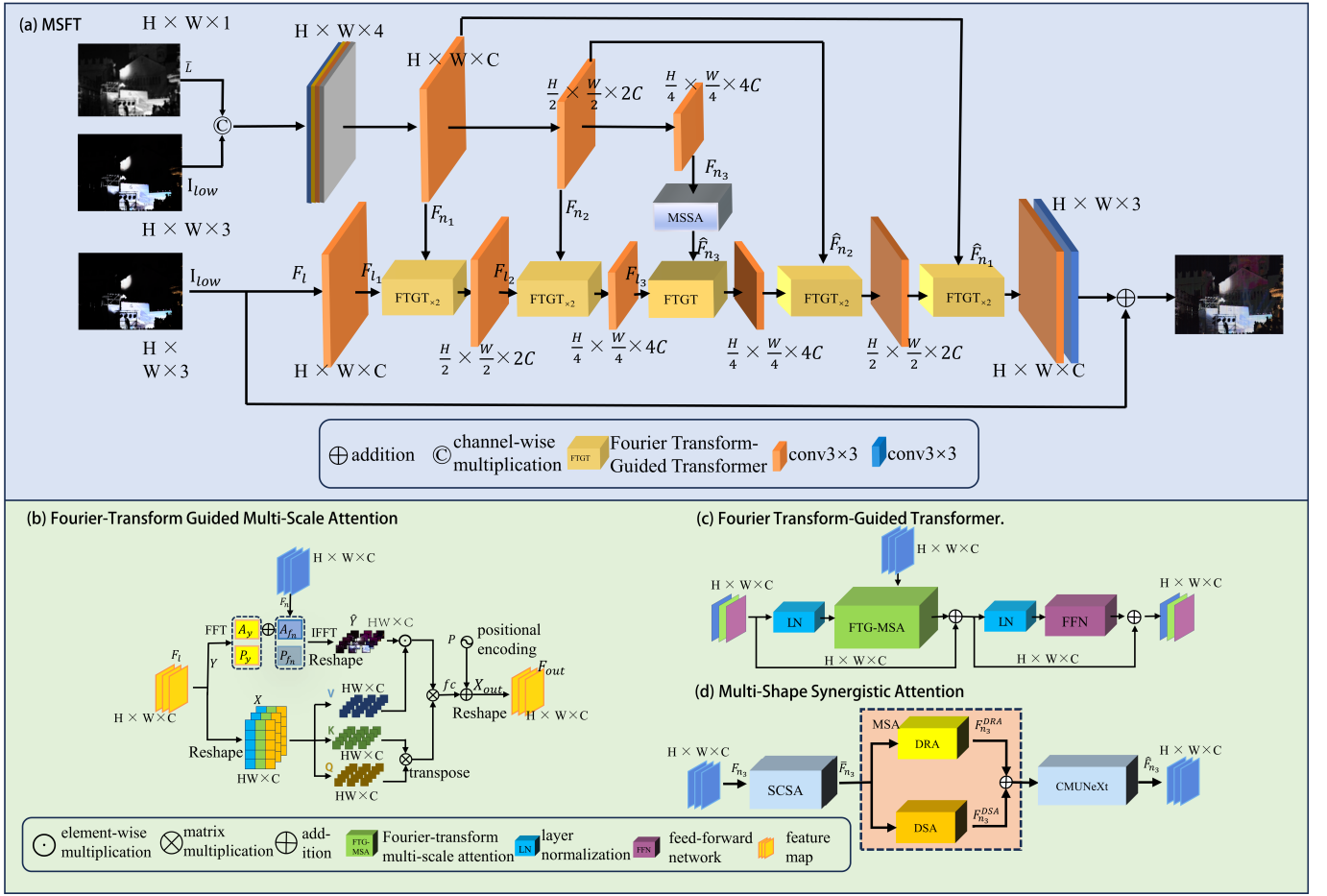}
\caption{(a) Overall structure of MSFT. (b) Schematic diagram of FTG-MSA. (c) Module diagram of FTGT. (d) Composition of MSSA.} 
\label{fig3.}
\end{figure}

Our overall architecture, as shown in Figure \ref{fig3.}, is a two-stage U-shaped framework. The detailed information is as follows: Figure \ref{fig3.}(a) presents MSFT, a U-shaped two-stage network with Transformer-based modules inserted at each scale. Figure \ref{fig3.}(b) illustrates FTG-MSA, which consists of two key components: \romannumeral1) a multi-head attention mechanism tailored for low-light images, and \romannumeral1) the fusion of frequency-domain amplitude spectra from a low-light image and a four-channel mapping image—this fused result is subsequently used as a guiding map.  Figure \ref{fig3.}(c) describes FTGT, comprising two Layer Normalization modules, one FTG-MSA module, and one Feed-Forward Network module. Figure \ref{fig3.}(d) details MSSA, which integrates three serially connected modules (SCSA, MSA, and CMUNeXt). MSA is constructed by parallel-connecting DSA and DRA. Next, we systematically elaborate on the complete construction of the MSFT architecture, focusing on three key components: 1) The Retinex decomposition performed in the original image space. 2) Mechanism of the Fourier-Transform Guided Multi-Scale Attention (FTG-MSA) module. 3) Structural design of the MSSA module.

\subsection{Multi-scale Attention Fourier-based Framework.}
\label{3.1}
According to Retinex theory, a low-light image $\textit{I}_{low} \in\mathbb{R}^{H\times W\times 3}$ can be decomposed into the illumination layer (\textit{L}) and the reflectance layer (\textit{R}) as
\begin{equation}
\label{Rtinex theory1}
I_{low}=L\odot R,
\end{equation}
where $\odot$ denotes the operation of the Hadamard product. Consistent with mainstream methods \citep{fu2023learning, wei2018deep, yi2023diff, zhang2021beyond}, we decompose the low illumination image in the image space and estimate the initial maps as follows, 
\begin{equation}
\label{Rtinex theory2}
\tilde{L}=\mathrm{max}_{c\in[0,C)}\textit{I}_{low}\in\mathbb{R}^{H\times W\times 1},
\end{equation}
where $C$ denotes the maximum number of channels, and concatenate the obtained illumination layer as the fourth channel with the three-channel RGB information of the low-light image $\text{I}_{low}$ to form the initial guiding input of the network as
\begin{equation}
\label{Rtinex theory3}
\textit{F}_n=\mathrm{Conv}(\textbf{\textit{I}}_{low},\tilde{L})\in\mathbb{R}^{H\times W\times 4}.
\end{equation}
 The purpose of this is to maintain content consistency between the guiding feature and the target feature in terms of brightness and contrast during the training process, to ensure the texture alignment of the feature. 

\subsection{Fourier Transform-Guided Transformer.}
\label{3.2}

Transformer-based deep learning approaches achieve remarkable image enhancement results, thanks to their superior capacity for capturing long-range dependencies. Retinexformer \citep{cai2023retinexformer} utilizes Transformer layers across multiple scales globally and fully exploits the potential of the attention mechanism by incorporating critical information. Refining the Retinexformer \citep{cai2023retinexformer}, we design a Fourier Transform-Guided Transformer (FTGT) to serve as a crucial restoration module.

\noindent{\bf{Network Structure.}}
The entire model framework is a three-scale U-shaped architecture \citep{10.1007/978-3-319-24574-4_28}, as shown in Figure \ref{fig3.}(a). The input of FTGT contains two parts: One part involves the information derived from four-channel mapping images after scale alignment and processing through a multi-shape attention mechanism $\textit{F}_n$, which serves as the crucial guiding information for low-light image enhancement. The other part is low-light images at different scales that require enhanced training $\textit{F}_l$. FTGT is a module that features dual inputs and a single output. In the encoding stage, $\textit{F}_l$ is subjected to a \textit{conv}3$\times$3 (which facilitates feature downscaling), followed by two FTGTs, an additional strided \textit{conv}3$\times$3, two additional FTGTs, and a final strided \textit{conv}3$\times$3 to generate hierarchical features $\textit{F}_{l_i} \in\mathbb{R}^{\frac{H}{2^{i}} \times\frac{W}{2^{i}}\times2^{i}C}$, where $C=32$ and $i = 0, 1, 2$. When $\textit{F}_l$ is fed as input to FTGT, the guidance information $\textit{F}_n$ also follows subsequently. $\textit{F}_n$ utilized undersampled convolution with the same parameters as $\textit{F}_l$ to obtain the corresponding rank features $\textit{F}_{n_i}$, where $i=0, 1, 2$. In the decoding stage, we directly feed the multi-scale features extracted during the encoding stage into their corresponding-scale decoder channels to guide the decoding process. This design eliminates the need for additional multi-scale feature extraction, thereby reducing computational redundancy and improving training efficiency. Finally, we utilize the Charbonnier loss to estimate the loss between the final output and the ground truth.

\noindent{\bf{FTG-MSA.}}
The input $\textit{F}_n$ of the FTG-MSA module in the FTGT is scale-aligned with the input $\textit{F}_l$ through a 3$\times$3 convolutional layer. For smaller scales, 3$\times$3 convolutional layers with a stride of 2 are employed to downscale the feature map $\textit{F}_l$ to match the corresponding spatial dimensions. To reduce computing costs, we only compute self-attention for $\textit{F}_l$.

Firstly, the input feature $\textit{F}_{l_i}\in\mathbb{R}^{H \times W\times C}$ is replicated into two branches. One branch denoted as $\textit{Y}\in\mathbb{R}^{H\times W\times C}$, when integrated with Fast Fourier Transform (FFT), has an amplitude characteristic described as
\begin{align}
\label{1}
(A_y, P_y)&=\mathrm{FFT}({\textit{Y}}), \nonumber    \\
(A_{f_n}, P_{f_n})&=\mathrm{FFT}({F}_n),
\end{align}
\noindent where $A_y$, $P_y$, $A_{f_n}$, and $P_{f_n}$ represent the amplitude and phase of $\textit{Y}$ and $\textit{F}_n\in\mathbb{R}^{H\times W\times C}$, respectively. $A_{f_n}$ incorporates luminance priors that align with the distribution of natural images. However, $P_{f_n}$ randomly extracts content from the sampled results, thus disrupting the generation of specific content. To effectively leverage the generated amplitude prior and to preserve the fundamental textural elements of the original image, we combine $A_y$ and $A_{f_n}$ to replace the amplitude. Next, we can derive the guiding factor and reshape it to $\widehat{\textit{Y}}\in\mathbb{R}^{HW\times C}$ through the application of the Inverse Fast Fourier Transform (IFFT),
\begin{equation}
\label{3}
\widehat{\textit{Y}}=\mathrm{Reshape}(\mathrm{IFFT}(A_y + A_{f_n},  P_n)),
\end{equation}
while the other branch is reshaped into tokens $\textit{X}\in\mathbb{R}^{HW\times C}$ to calculate self-attention. Then \textit{X} is split into \textit{k} heads,
\begin{equation}
\label{4}
X=[{{X}}_1, {{X}}_2, \cdots, {{X}}_{k}],
\end{equation}
\noindent where $\textit{X}_i \in \mathbb{R}^{HW \times d_k}$, $d_k= {C}/{k}$, and $i=1, 2, \cdots, k$. In Figure ~\ref{fig3.}(b), we describe the scenario in which $k=1$. For each $head_i$, three fully connected $(fc)$ layers without $bias$ are used to linearly project $\textit{X}_i$ into $query$ elements $Q_i \in \mathbb{R}^{HW \times d_k}$, $key$ elements $K_i \in \mathbb{R}^{HW \times d_k}$, and $value$ elements $V_i \in \mathbb{R}^{HW \times d_k}$ as
\begin{equation}
\label{5}
{Q}_i={{X}}_i{W}_{{Q}_i}^T,\
{K}_i={{X}}_i{W}_{{K}_i}^T,\
{V}_i={{X}}_i{W}_{{V}_i}^T,
\end{equation}
\noindent where $W_{Q_i}$, $W_{K_i}$, and $W_{V_i} \in \mathbb{R}^{d_k \times d_k}$
represent the learnable parameters of the $FC$ layers, and $T$ denotes the matrix transpose. As mentioned above, $\widehat{Y}$, which incorporates the amplitude of $F_n$, contains both the brightness information of the four-channel feature brightness map necessary for training and the texture characteristics of the original low-light image. Thus, we split $\widehat{Y}$ into $k$ heads to align with the shape of $X$,
\begin{equation}
\label{6}
{\widehat{Y}}=[{\widehat{Y}}_1, {\widehat{Y}}_2, \cdots, {\widehat{Y}}_k],
\end{equation}
\noindent where $\widehat{Y}_i \in \mathbb{R}^{HW \times d_k}$, $d_k= {C}/{k}$, and $i=1, 2, \cdots, k$. Then the self-attention for each $head_i$ is formulated as
\begin{equation}
\label{7}
\mathrm{Attention}({Q}_i,{K}_i, {V}_i, \widehat{\textbf{Y}}_i) = (\widehat{\textbf{Y}}_i\odot{V}_i)\mathrm{softmax}(\frac{{K}_i^T{Q}_i}{\sqrt{d_k}}).
\end{equation}
\noindent Next, the heads of $k$ are then concatenated and pass through a fully connected layer $(fc)$. This is followed by the addition of a positional encoding $\textit{P}\in\mathbb{R}^{HW \times C}$ to produce the output tokens $\textit{X}_{out}\in\mathbb{R}^{HW\times C}$. Finally, we reshape $\textit{X}_{out}$ to the estimated feature $\textit{F}_{out}\in\mathbb{R}^{H\times W\times C}$.

\subsection{Multi-Shape Synergistic Attention}
\label{3.3}

To better extract multi-scale feature representation capability from normal light images, we propose an MSSA module to accommodate our multi-scale feature extraction, as shown in Figure \ref{fig3.}(d), which involves three parts: an MSA \citep{10713101} module, an SCSA module, and a dynamic filter network. Firstly, the MSA module plays a crucial role in enhancing the feature representation capabilities of image restoration models across multiple scales and shapes, while maintaining low computational complexity. Specifically, the MSA module integrates the advantages of self-attention mechanisms with convolutional operations, exhibiting content-awareness while preserving the efficiency inherent to convolutional processes. In addition, we select SCSA connected in series before MSA to adaptively weigh the importance of different channels and augment critical spatial information. Finally, we leverage dynamic network filters to augment the comprehensive receptive field. The highest-dimensional paired normal light images are normalized to obtain $\textit{F}_{n_3}\in\mathbb{R}^{H\times W\times C}$ after a block $conv3\times3$ to align the highest-dimensional low-light images. Subsequently, $\textit{F}_{n_{3}}$ proceeds through the SCSA module as
\begin{equation}
\label{3.2.1}
\bar{F}_{n_{3}}=\Phi_{\mathrm{SCSA}}(F_{n_{3}}),
\end{equation}
\noindent where the SCSA module does not alter the channel dimensions of the training data. The MSA module consists of Dilated Square Attention (DSA) and Dilated Rectangle Attention (DRA) in parallel, aggregating information in  square and rectangular regions respectively, to realize feature learning at different scales.

\noindent \textbf{DSA:} DSA generates attentional weights that adapt to input features through the convolutional layer, and uses a hyperbolic tangent function (Tanh) instead of the original Softmax function to avoid the limitations of low-pass filters and enhance the high-frequency components. The magnitude of the high-frequency components is further increased by modifying the spectral weight to the filter, so more attention can be paid to the high-frequency information in the image during the polymerization process. For given $\bar{F}_{n_3}\in\mathbb{R}^{H \times W\times C}$, the output features can be formally obtained by
\begin{eqnarray}
\label{3.2.2}
A^{DSA}&=&\mathrm{Tanh}(\mathrm{Conv}_{1\times1}(\mathrm{GAP}(\mathrm{Conv}_{3\times3}(\bar{F}_{n_{3}})))), \nonumber \\
A_{h}^{DSA}&=&A^{DSA}-A_{l}^{DSA},\\
\hat{A}^{DSA}&=&A_{l}^{DSA}+WA_{h}^{DSA}, \nonumber
\end{eqnarray}
\noindent where $A_{l}^{DSA}=\frac{1}{K^2}E$, $A^{DSA}, E\in\mathbb{R}^{G\times K\times K}$; $G$ is the number of feature groups and $K^2$ is the absolute region size for integration, GAP refers to global average pooling, $W$ denotes the learnable parameters directly optimized by backpropagation and initialized as \textbf{1}. Finally, we apply the resultant spatial-channel-synergistic-attention weights to the input via SCSA operations, the output can be obtained by
\begin{equation}
\begin{aligned}
\hat{F}_{n_{3}, (g,h,w)}^{DSA}=&\sum^{K-1}_{i=0}\sum^{K-1}_{j=0}\bar{F}_{n_{3},[g,(h-\lfloor\frac{K}{2}\rfloor+i)d,(w-\lfloor\frac{K}{2}\rfloor+j)d]}\hat{A}^{DSA}_{g,i,j} \\
&+\bar{F}_{n_{3},(g,h,w)},
\end{aligned}
\end{equation}
\noindent where $g$, $h$, and $w$ are the indices of the group, height, and width, respectively, and $d$ is the dilation rate.

\noindent \textbf{DRA:} DRA conducts information aggregation in a rectangular manner to achieve multi-shape feature representation capabilities. Analogous to DSA, the DRA module employs convolutional networks to generate initial attention weights and re-evaluates the significance of high-frequency components via filter modulation with rectangular shapes. Similarly, for the input $\bar{F}_{n_3}\in\mathbb{R}^{H \times W\times C}$, the output feature of the horizontal unit is defined as
\begin{eqnarray}
\label{3.2.6}
\hat{F}_{n_{3},(g,h,w)}^H&=&\sum^{K-1}_{j=0}\bar{F}_{n_{3},[g,h,(w-\lfloor\frac{K}{2}\rfloor+j)d]}\widetilde{A}^H_{g,j}+\bar{F}_{n_{3},(g,h,w)}, \nonumber \\
\widetilde{A}^H&=&\mathrm{RFM}(A^H),\\
A^H&=&\mathrm{Tanh}(\mathrm{Conv}_{1\times 1}(\mathrm{GAP}(\bar{F}_{n_{3}})))\in\mathbb{R}^{G\times K}, \nonumber
\end{eqnarray}
\noindent where $A^{H}$ denotes the attention map of the horizontal unit in DRA; RFM refers to the application of FM to attention weights that are structured in a rectangular format. The derivation process of the horizontal unit can be written as 
\begin{equation}
\hat{F}^H=\mathcal{H}(\bar{F}).
\end{equation}
Then, using the output of the horizontal unit as input, the vertical unit generates its output,
\begin{equation}
\hat{F}^V=\mathcal{V}(\hat{F}^H).
\end{equation}
Finally, the result of DRA can be formally expressed as 
\begin{equation}
{F}^{MSA}_{n_3}=\hat{F}^V+\bar{F}_{n_3}.
\end{equation}
To integrate global spatial information and channel information extracted through convolution, we introduce the CMUNeXt \citep{10812276} module after the MSA module. This module is capable of extracting global information across all channels simultaneously, while effectively reducing network parameters and computational costs. The CMUNeXt block is defined as  
\begin{eqnarray}
\label{CMUNeXt}
F'_{n_3}&=&\mathrm{BN(GELU(DepthwiseConv}(F^{MSA}_{n_3})))+F^{MSA}_{n_3}, \nonumber \\
F''_{n_3}&=&\mathrm{BN(GELU(PointwiseConv}
(F'_{n_3}))), \\
\hat{F}_{n_3}&=&\mathrm{BN(GELU(PointwiseConv}
(F''_{n_3}))), \nonumber
\end{eqnarray}
where $\hat{F}_{n_3}$ is the final extracted feature map, BN, DepthwiseConv, GELU, and PointwiseConv refer to batch normalization, pointwise convolution, Gaussian Error Linear Unit, and depthwise separable convolution, respectively.

\section{Experiments}
\subsection{ Datasets and Implementation Details}

We evaluated the performance of our model on the LOL (v1 \citep{wei2018deep}, v2 Real and v2 Synthetic \citep{9328179}), SID \citep{9009494}, SMID \citep{8578445}, and SDSD (indoor and outdoor) datasets \citep{9710730}. The key characteristics of each dataset are summarized in Table \ref{TabB}

\hyphenation{
    ar-chI-tec-ture en-hance-ment de-com-po-si-tion re-con-struc-tion
    il-lu-mi-na-tion pre-dic-tion pa-ra-me-ters gen-er-al-i-za-bil-i-ty
}
\newcolumntype{S}{>{\RaggedRight\arraybackslash\hyphenpenalty=5000\exhyphenpenalty=500}>{\hsize=0.8\hsize}X}    
\newcolumntype{M}{>{\RaggedRight\arraybackslash\hyphenpenalty=5000\exhyphenpenalty=500}>{\hsize=1.0\hsize}X}    
\newcolumntype{R}{>{\RaggedRight\arraybackslash\hyphenpenalty=5000\exhyphenpenalty=500}>{\hsize=0.6\hsize}X}    
\newcolumntype{I}{>{\RaggedRight\arraybackslash\hyphenpenalty=5000\exhyphenpenalty=500}>{\hsize=1.1\hsize}X}    
\newcolumntype{Z}{>{\RaggedRight\arraybackslash\hyphenpenalty=5000\exhyphenpenalty=500}>{\hsize=1.5\hsize}X}    

\begin{table*}[!t]
\centering
\fontsize{8}{9}\selectfont
\small
\renewcommand{\arraystretch}{1.4}  
\caption{{Comparison of Related Low-Light Enhancement Datasets}}
\label{tab:low_light_enhancement_datasets}
\begin{tabularx}{\textwidth}{S M R I Z}
\toprule
{Datasets}        & {Captured scene}                                                                 & {Resolution}                                                                 & {Implementation details}                                                              & {Summary}                                                              \\ 
\midrule
{LOLv1 \citep{wei2018deep}}       & {real indoor scenes}                                                                   & \multirow{3}{*}{{400×600}}                               & {Total number 500, 485 for training and 15 for testing.}                                          & \multirow{3}{*}[-3pt]{\parbox{\dimexpr0.28\textwidth-2\tabcolsep\relax}{{LOL contains inherent noise from the capture process, making the datasets a benchmark for LLIE, detail restoration, and denoising.}}} \\
\addlinespace
{LOLv2-real \citep{9328179}} & {real outdoor scenes}                        &                                & {Total number 789, 689 for training and 100 for testing.}                        &  \\
\addlinespace
{LOLv2-syn \citep{9328179}}  & {synthetic outdoor scenes}                                       & & {Total number 1000, 900 for training and 100 for testing.}                  &  \\
\addlinespace
{SID \citep{9009494}}    & {real indoor and outdoor scenes}                                        & {768×1280}                                                                  & {2099 pairs for training and 589 pairs for testing.}                            & {The SID dataset captured by the Sony 7S $\alpha$II camera contains indoor and outdoor images, totaling 2697 short-/long-exposure RAW image pairs} \\
\addlinespace
{SMID \citep{8578445}}     & {static indoor scenes and dynamic outdoor scenes}           & {3672×5496}            & {15,763 pairs for training and 5046 for testing.}                    & {The SMID benchmark comprises 20,809 short-/long-exposure RAW image pairs, including static indoor scenes and dynamic outdoor scenes.} \\
{SDSD-indoor \citep{9710730}} & {static indoor scenes} & {512 × 960} & {Total number 68, 62 for training and 6 for testing.} & \multirow{2}{*}[-14pt]{\parbox{\dimexpr0.28\textwidth-2\tabcolsep\relax}{{The SDSD dataset captured by a Canon EOS 6D Mark II with an ND filter is collected as dynamic video pairs containing low-light and normal-light videos.}}} \\
\addlinespace 
{SDSD-outdoor \citep{9710730}} & {static outdoor scenes} & {1080×1920} & {Total number 126, 116 for training and 10 for testing.} & \\
 \addlinespace
  \addlinespace
\bottomrule
\end{tabularx}
\label{TabB}
\end{table*}

\noindent{\bf{LOL}}. The LOL dataset primarily comprises indoor scene images with a resolution of 400$\times$600. These images contain inherent noise from the capture process, making the dataset a benchmark for tasks like low-light image enhancement, detail restoration, and denoising. In this experiment, the LOL dataset is divided into two parts: v1 and v2. LOL-v2 is divided into real and synthetic subsets. The training and testing sets are split as follows: 485 to 15 for LOL-v1, 689 to 100 for LOL-v2-real, and 900 to 100 for LOL-v2-synthetic.

\noindent{\bf{SID}}. The SID dataset captured by the Sony 7S $\alpha$II camera contains indoor and outdoor images, totaling 2697 short-/long-exposure RAW image pairs. The illuminance of outdoor scenes ranges from 0.2 to 5 lux, while that of indoor scenes is between 0.03 and 0.3 lux. The exposure time of short-exposure images is set to 1/30 to 1/10 seconds, and the corresponding reference long-exposure images have an exposure time 100 to 300 times longer, ranging from 10 to 30 seconds. Low-/normal-light RGB images are obtained through the same in-camera signal processing as SID's RAW-to-RGB conversion. 2099 pairs are used for training and 598 for testing.

\noindent{\bf{SMID}}. The SMID benchmark comprises 20,809 short-/long-exposure RAW image pairs, including static indoor scenes and dynamic outdoor scenes. The resolution of the raw data is 3672$\times$5496, and the brightness levels in the dataset range from 0.5 to 5 lux. Additionally, the RAW data were converted into low-/normal-light RGB image pairs. Of these, 15,763 pairs were utilized for training, while the remaining pairs were reserved for testing.

\noindent{\bf{SDSD}}. The SDSD dataset is collected as dynamic video pairs containing low-light and normal-light videos. This dataset consists of indoor and outdoor subsets. We used the static version of SDSD, captured by a Canon EOS 6D Mark II with an ND filter. The resolution of indoor and outdoor subsets are 512 $\times$ 960 and 1080 $\times$ 1920 respectively. For training and testing, we use 62 to 6 low-/normal-light video pairs for SDSD-indoor and 116 to 10 pairs for SDSD-outdoor.

\noindent{\bf{Implementation Details.}} We implement the proposed model by the PyTorch deep learning framework. All experiments were conducted on a Linux platform equipped with an Intel i9-14900HX CPU and an NVIDIA RTX 3090 GPU with 24 GB of memory. The model is trained with Adam Optimizer ($\beta_1=0.9$ and $\beta_2=0.999$) for $1.5\times 10^5$ iterations. The learning rate is initially set to $2\times 10^{-6}$ and then steadily decreased to $1\times 10^{-6}$ by the cosine annealing scheme during the training process. Patches of size 128×128 are randomly cropped from pairs of low-light and normal-light images to serve as training samples. The batch size is set to 8. To enhance the training data, we apply random rotation and flipping as augmentations. The primary objective of the training process is to minimize the mean absolute error (MAE) between the enhanced images and their corresponding ground truth. For evaluation purposes, we use Peak Signal-to-Noise Ratio (PSNR) and Structural Similarity Index Measure (SSIM) \citep{1284395} as metrics. 

\begin{figure}[htbp]
\centering
\includegraphics[width=1.0\textwidth]{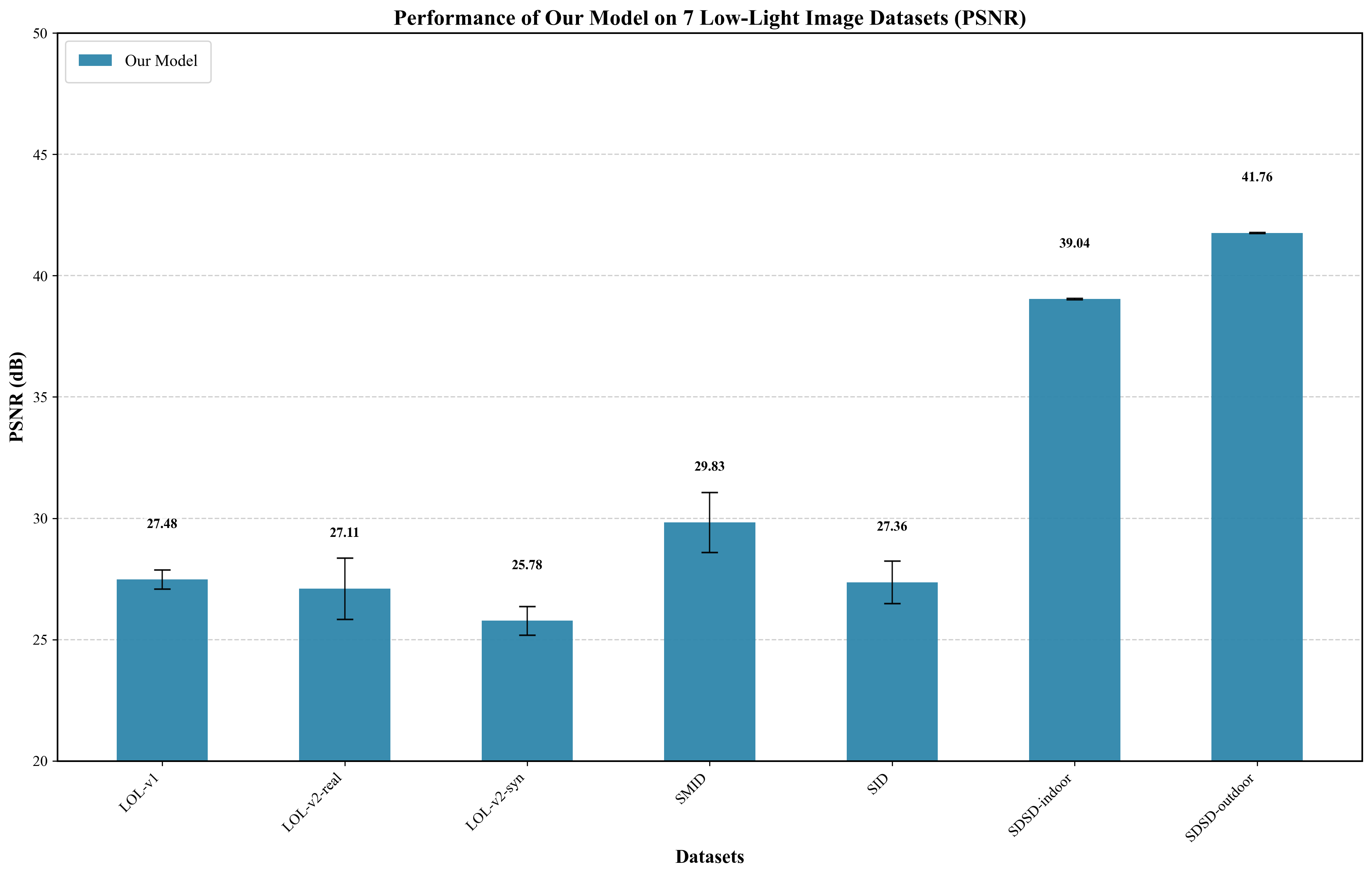}
\caption{visualization results of repeated runs for each dataset.}
\label{run}
\end{figure}

\noindent{\bf{Statistical analysis of test results.}}
We executed 12 independent runs of the MSFT algorithm on 7 datasets. From these, the model exhibiting the highest PSNR on the validation set was selected for final evaluation. Its performance was then assessed on the separate test sets. Furthermore, a statistical analysis was conducted on the convergence results from all twelve runs to verify the consistency and stability of our experimental outcomes. The statistical data results and the visualization results are respectively shown in Table  \ref{Tab0} and Figure \ref{run}. The data results delivered optimal and highly stable results on the SDSD-outdoor and SDSD-indoor datasets, with PSNRs of 41.76±0.03 and 39.04±0.02 and very narrow confidence intervals, demonstrating strong robustness in these scenarios. In contrast, performance dropped significantly on the LOL-v2-syn dataset, averaging only 25.78±0.66, which underscores the domain shift challenge between synthetic and real data. The model also exhibited considerable instability on the SMID and LOL-v2-real datasets, where standard deviations exceeded 1.38 and confidence intervals were broader, indicating limited robustness in complex real-world settings. Further comparison of confidence intervals revealed consistent performance between the LOL-v1 and SID datasets, suggesting good cross-dataset generalization. However, a noticeable performance gap between LOL-v2-real and LOL-v1 highlights the impact of internal distribution shifts even within real data. In summary, while the model performs excellently on certain datasets, its generalization capability and reliability in more challenging real-world environments remain areas for future improvement.

\subsection{Low-light Image Enhancement}

\begin{figure}[htbp]
\centering
\includegraphics[width=0.9\textwidth]{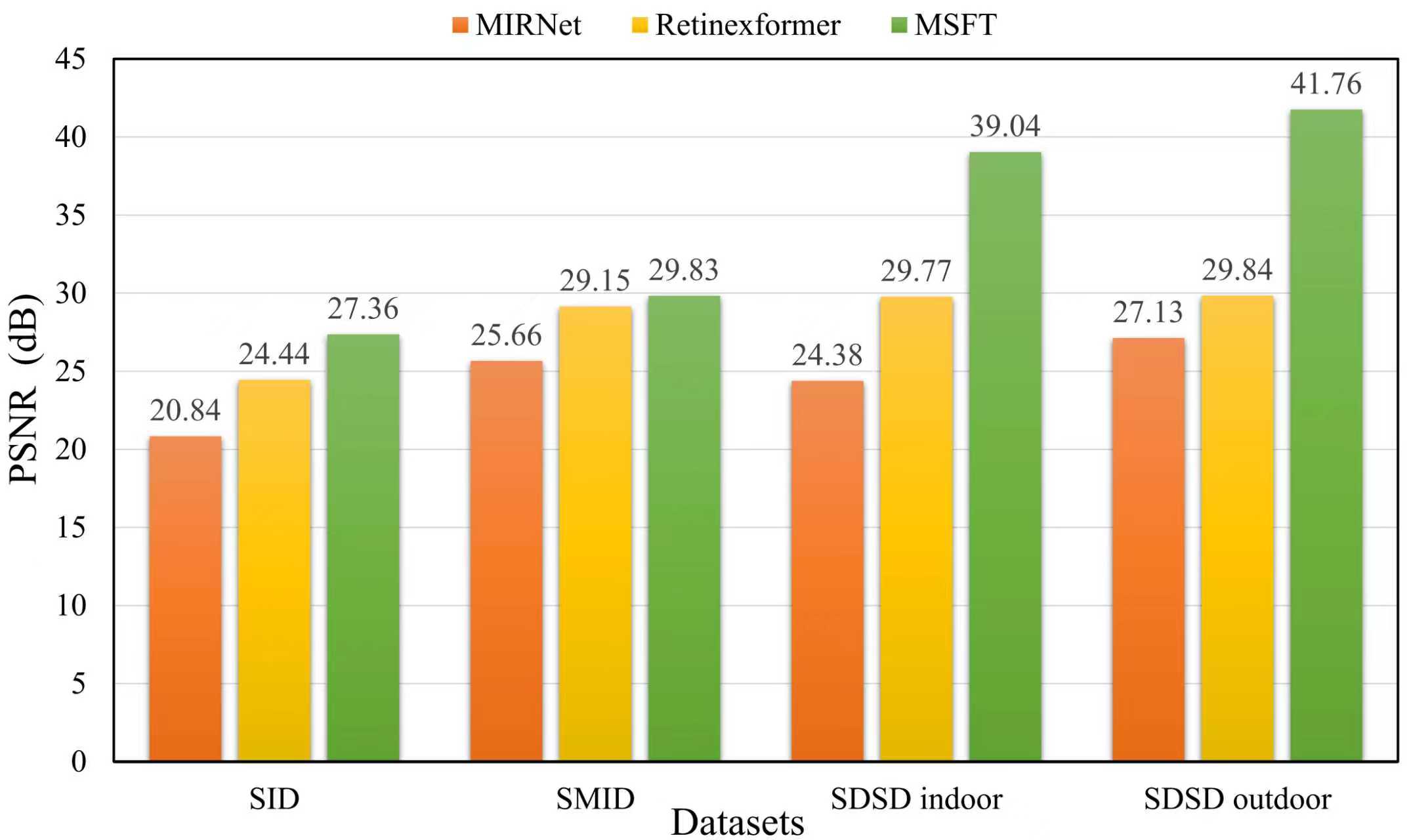}\caption{Comparison charts on four datasets.}
\label{fig1}
\end{figure}

\noindent{\bf{Baselines.}}
As shown in Table \ref{Tab1}, we compare our method with existing LLIE methods, which are divided into three categories, on the four datasets mentioned above: 1) unsupervised methods: SNR-Net \citep{xu2022snr}, Retinexformer \citep{cai2023retinexformer}, EnGAN \citep{jiang2021enlightengan}, and RUAS \citep{DBLP:conf/cvpr/Liu0Z0L21}, 2) semi-supervised methods: 3DLUT \citep{journals/pami/ZengCLCZ22} and DRBN \citep{9369069}, 3) superivised methods: SID \citep{9009494}, DeepUPE \citep{8953588}, DeepLPF \citep{9156461}, IPT \citep{9577359}, Uformer \citep{9878729}, RetinexNet \citep{LI2024103948}, Sparse \citep{9328179}, FIDE \citep{9156446}, KinD \citep{zhang2019kindlingdarknesspracticallowlight}, Restormer \citep{9878962}, and MIRNet \citep{9756908}.  MSFT outperforms all supervised, unsupervised, and semi-supervised models. As shown in Figure \ref{fig1}, in terms of PSNR values, MSFT, significantly outperforms Retinexformer and MIRNet on four LLIE benchmarks: SID, SMID, SDSD-indoor/outdoor.

\begin{table}[t]
\renewcommand{\arraystretch}{1.25}
    \caption{ The statistical data results of repeated runs for each dataset.}
    \resizebox{1.0\textwidth}{!}
    {    
    \begin{tabular}{cccccccc}
    \toprule[0.5mm]
    \textbf{Datasets}                 & LOL-v1            & LOL-v2-real        & LOL-v2-syn         & SMID              & SID                & SDSD-indoor        & SDSD-outdoor       \\ \hline
    \textbf{our model (mean ± std)}   & 27.48±0.44         & 27.11±1.41         & 25.78±0.66        & 29.83±1.38         & 27.36±0.97         & 39.04±0.02         & 41.76±0.03         \\
    \textbf{99\% confidence interval} & [27.09, 27.87] & {[}25.84, 28.37{]} & {[}25.19, 26.37{]} & {[}28.60, 31.06{]} & {[}26.49, 28.24{]} & {[}39.02, 39.06{]} & {[}41.74, 41.79{]} \\ 
    \bottomrule[0.5mm]
    \label{Tab0}
    \end{tabular}
    }   
\end{table}

\begin{table}[htbp]
    \centering
    \caption{Quantitative comparisons with three
    categories of state-of-the-art methods on the seven datasets. The best and the second-best evaluation values are marked in red and blue fonts, respectively. MSFT significantly outperforms other SOTA algorithms. The evaluation metrics are PSNR$\uparrow$ and SSIM$\uparrow$.}
    \renewcommand{\arraystretch}{1.25}
    \resizebox{1.0\textwidth}{!}
        {
        \begin{tabular}{cclcccccccccccccc}
        \toprule[0.5mm]
            \multirow{2}{*}{Type} & \multicolumn{2}{c}{\multirow{2}{*}{Methods}}       & \multicolumn{2}{c}{LOL-v1} & \multicolumn{2}{c}{LOL-v2-real} & \multicolumn{2}{c}{LOL-v2-syn} & \multicolumn{2}{c}{SMID} & \multicolumn{2}{c}{SID} & \multicolumn{2}{c}{SDSD-indoor} & \multicolumn{2}{c}{SDSD-outdoor} \\ \cline{4-17} 
             & \multicolumn{2}{c}{}              & PSNR         & SSIM         & PSNR            & SSIM & PSNR           & SSIM           & PSNR        & SSIM        & PSNR        & SSIM       & PSNR            & SSIM           & PSNR            & SSIM            \\ \midrule[0.5mm]
            \multirow{4}{*}{UL}   & \multicolumn{2}{c}{EnGAN}         & 17.48        & 0.650         & 18.23           & 0.617          & 16.57          & 0.734          & 22.62       & 0.674       & 17.23       & 0.543      & 20.02           & 0.604          & 20.10            & 0.616           \\
              & \multicolumn{2}{c}{RUAS}          & 18.23        & 0.720         & 18.37           & 0.723          & 16.55          & 0.652          & 25.88       & 0.744       & 18.44       & 0.581      & 23.17           & 0.696          & 23.84           & 0.743           \\
               & \multicolumn{2}{c}{SNR-Net}       & 24.61        & 0.842        & 21.48           & \color[HTML]{3166FF}0.849          & 24.14          & \color[HTML]{3166FF}{0.928}          & 28.49       & 0.805       & 22.87       & 0.625      & 29.44           & 0.894          & 28.66           & 0.866           \\
                & \multicolumn{2}{c}{Retinexformer} & \color[HTML]{3166FF}{25.16}        &\color[HTML]{3166FF}{ 0.845}        & \color[HTML]{3166FF}{22.80}            & {0.840}           & \color[HTML]{3166FF}{25.67}          & \color[HTML]{FE0000}{0.930}           & \color[HTML]{3166FF}{29.15}       & \color[HTML]{3166FF}{0.815}       & \color[HTML]{3166FF}{24.44}    & \color[HTML]{3166FF}{0.680}       & \color[HTML]{3166FF}{29.77}           & \color[HTML]{3166FF}{0.896}          & \color[HTML]{3166FF}{29.84}           & \color[HTML]{3166FF}{0.877}           \\ \hline
            \multirow{2}{*}{SSL}  & \multicolumn{2}{c}{3DLUT}         & 14.35        & 0.445        & 17.59           & 0.721          & 18.04          & 0.800            & 23.86       & 0.678       & 20.11       & 0.592      & 21.66           & 0.655          & 21.89           & 0.649           \\
             & \multicolumn{2}{c}{DRBN}          & 20.13        & 0.830         & 20.29           & 0.831          & 23.22          & 0.927          & 26.60        & 0.781       & 19.02       & 0.577      & 24.08           & 0.868          & 25.77           & 0.841           \\ \hline
            \multirow{12}{*}{SL}  & \multicolumn{2}{c}{SID}           & 14.35        & 0.436        & 13.24           & 0.442          & 15.04          & 0.610           & 24.78       & 0.718       & 16.97       & 0.591      & 23.29           & 0.703          & 24.90            & 0.693           \\
             & \multicolumn{2}{c}{DeepUPF}       & 14.38        & 0.446        & 13.27           & 0.452          & 15.08          & 0.623          & 23.91       & 0.690        & 17.01       & 0.604       & 21.70            & 0.662          & 21.94           & 0.698           \\
             & \multicolumn{2}{c}{DeepLPE}       & 15.28        & 0.473        & 14.10            & 0.480  & 16.02          & 0.687          &24.36        & 0.688         & 18.07       & 0.600      & 22.21           & 0.664          & 22.76           & 0.658           \\
              & \multicolumn{2}{c}{IPT}           & 16.27        & 0.504        & 19.80            & 0.813          & 18.30           & 0.811          & 27.03       & 0.783       & 20.53       & 0.561      & 26.11           & 0.831          & 27.55           & 0.850            \\
              & \multicolumn{2}{c}{Uformer}       & 16.36        & 0.771        & 18.82           & 0.771          & 19.66          & 0.871          & 27.20        & 0.792       & 18.54       & 0.577      & 23.17           & 0.859          & 23.85           & 0.748           \\
              & \multicolumn{2}{c}{RetinexNet}    & 16.77        & 0.56         & 15.47           & 0.567          & 17.13          & 0.798          & 22.83       & 0.684       & 16.48       & 0.578      & 20.84           & 0.617          & 20.96           & 0.629           \\
               & \multicolumn{2}{c}{Sparse}        & 17.20        & 0.640         & 20.06           & 0.816          & 22.05          & 0.905          & 25.48       & 0.766       & 18.68     & 0.606      & 23.25           & 0.863          & 25.28           & 0.804           \\
                & \multicolumn{2}{c}{FIDE}          & 18.27        & 0.665        & 16.85           & 0.678          & 15.20         & 0.612                 & 24.42       & 0.692         & 18.34       & 0.578    & 22.41           & 0.659          & 22.20            & 0.629           \\
                & \multicolumn{2}{c}{KinD}          & 20.86        & 0.790         & 14.74           & 0.641          & 13.29          & 0.578          & 22.18       & 0.634       & 18.02       & 0.583      & 21.95           & 0.672          & 21.97           & 0.654           \\
                 & \multicolumn{2}{c}{Restormer}     & 22.43        & 0.823        & 19.94           & 0.827          & 21.41          & 0.830            & 26.97          & 0.758     & 22.27       & 0.649       & 25.67           & 0.827          & 24.79           & 0.802           \\
                & \multicolumn{2}{c}{MIRNet}        & 24.14        & 0.830         & 20.02           & 0.820         & 21.94          & 0.876          & 25.66       & 0.762       & 20.84       & 0.605      & 24.38           & 0.864          & 27.13           & 0.837           \\
                \midrule[0.5mm]
                & \multicolumn{2}{c}{MSFT}          & \color[HTML]{FE0000}{27.48}        & \color[HTML]{FE0000}{0.868}        & \color[HTML]{FE0000}{27.11}           & \color[HTML]{FE0000}{0.859}          & \color[HTML]{FE0000}{25.78}          & 0.902          & \color[HTML]{FE0000}{29.83}        & \color[HTML]{FE0000}{0.866}       & \color[HTML]{FE0000}{27.36}       & \color[HTML]{FE0000}{0.834}      & \color[HTML]{FE0000}{39.04}           & \color[HTML]{FE0000}{0.974}         & \color[HTML]{FE0000}{41.76}           & \color[HTML]{FE0000}{0.988}          \\
                \bottomrule[0.5mm]
            \end{tabular}
            }
    \label{Tab1}
\end{table}

\noindent{\bf{Quantitative Results.}}
We conduct a quantitative comparison of the proposed method against a wide variety of state-of-the-art enhancement algorithms, as presented in Table \ref{Tab1}. We simplify the comparative analysis of traditional LLIE methods.

Compared to methods from unsupervised deep learning, our method outperforms the state-of-the-art Retinexformer by achieving significant PSNR improvements of  2.32, 4.29, 0.11, 0.68, 2.92, 9.27, and 11.92 dB on seven benchmark datasets, including LOL-v1, LOL-v2-real, LOL-v2-synthetic, SID, SMID, SDSD-indoor, and SDSD-outdoor, respectively. Especially on the SDSD indoor and outdoor datasets, the SSIM of MIRNet and IPI are increased by 12.7 $\%$ and 16.24$\%$, respectively. 

Compared with SOTA supervised deep learning methods, MSFT achieves 3.34, 7.03, 3.73, 2.63, 5.09, 12.93, and 16.48 dB improvements on the seven benchmarks in Table \ref{Tab1}. In particular, we notice that the improvements on the SDSD indoor/outdoor dataset are all over \textbf{9} dB, and their estimated SSIM values are all above \textbf{0.95}.
The quantitative analysis mentioned above suggests that MSFT has a strong competitive advantage compared with existing methods.

\begin{figure}[t]
\centering
\includegraphics[width=1.0\textwidth]{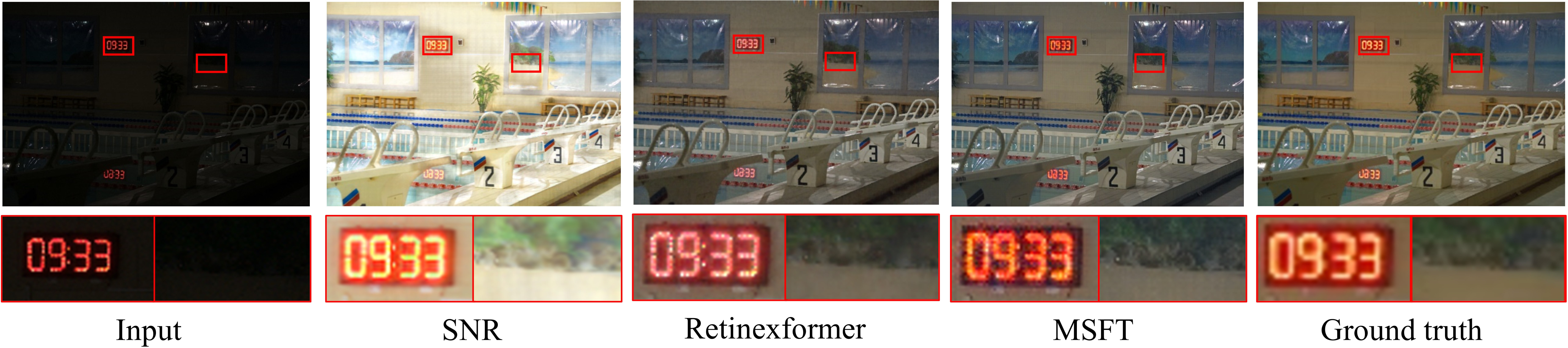}
\caption{Visualization of qualitative comparisons with SNR-Net, and Retinexformer on the LOLv1 dataset.}
\label{fig4}
\end{figure}
\begin{figure}[!t]
\centering
\includegraphics[width=1.0\textwidth]{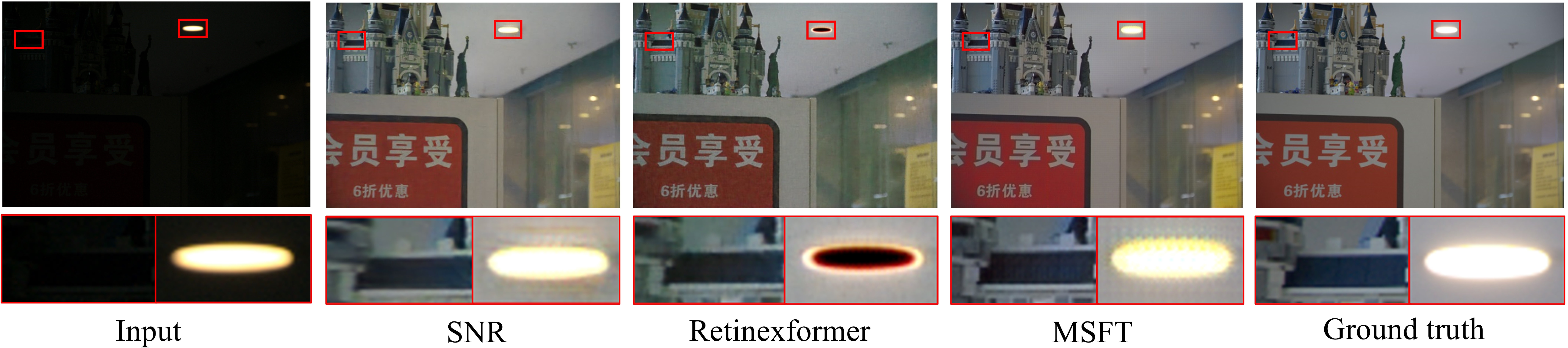}
\caption{Visualization of qualitative comparisons with SNR-Net, and Retinexformer on the LOLv2 synthetic dataset.}
\label{fig5}
\end{figure}
\begin{figure}[htbp]
\centering
\includegraphics[width=1.0\textwidth]{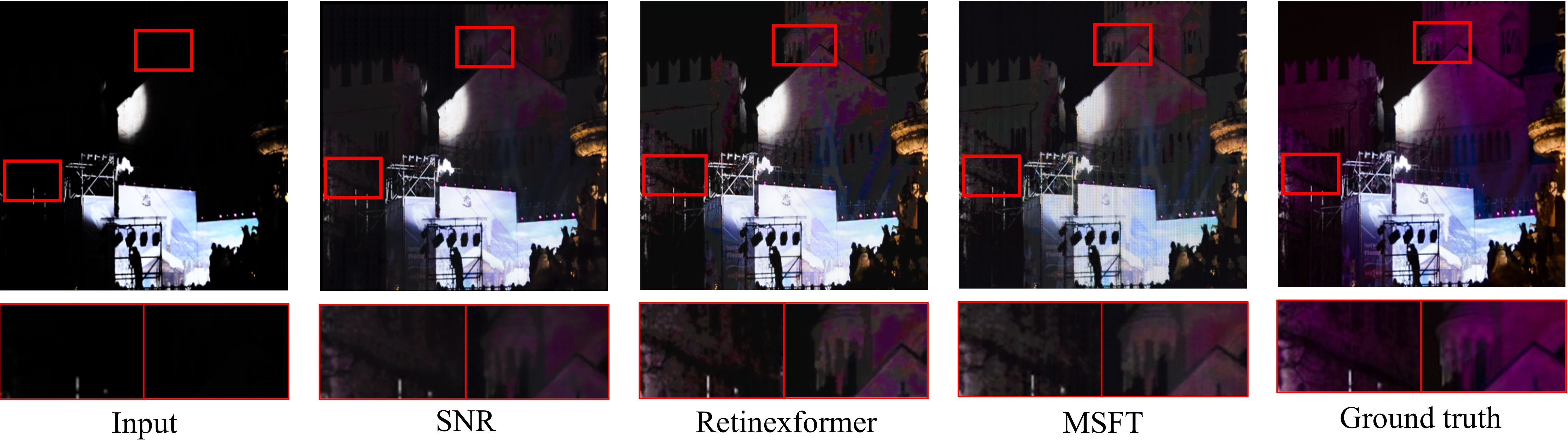}
\caption{Visualization of qualitative comparisons with SNR-Net,  and Retinexformer on the LOLv2 real dataset.}
\label{fig6}
\end{figure}

\begin{figure}[!t]
\centering
\includegraphics[width=1.0\textwidth]{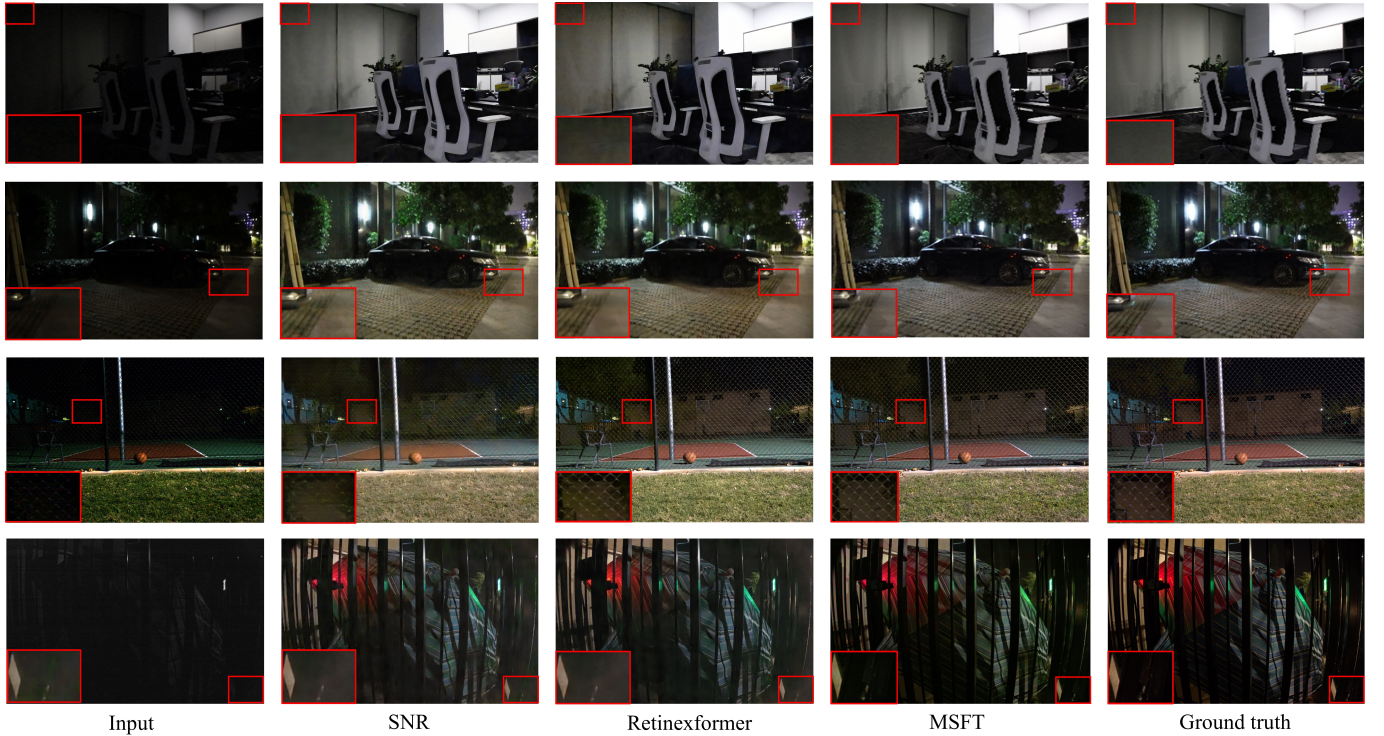}
\caption{Visualization of qualitative comparisons with SNR-Net,  and Retinexformer on the SDSD-indoor (top row), SDSD-outdoor (2nd row), SID (3rd row), and SMID (4th row) dataset.}
\label{fig7}
\end{figure}

\noindent{\bf{Params and FLOPs of Algorithm  Analysis}}
We evaluated the size and performance of the model by calculating the computational cost and parameter quantity of the model with a fixed image size of $1\times3\times256\times256$, and comparing it with four SOTA enhancement methods on the LOL-v1 datasets in Table \ref{tab5}. At the same time, we evaluated the size of each unit in the MSSA module to verify its lightweight nature during computation. The result analysis shows that MSFT has the highest PSNR of 27.48 decibels, and its parameter quantity is 1.25 million and the computational quantity is 18.07 billion floating-point operations, which are relatively low. Thus, its overall performance is the best. DeepLPF has the smallest computational quantity of 5.86 billion floating-point operations, but its PSNR is only 15.28 decibels, with limited performance. SNR-Net has the largest parameter quantity of 4.01 million and the highest computational quantity, with a PSNR of 24.61 decibels. Retinexformer is relatively balanced in all indicators, with a PSNR of 22.43 decibels. DeepUPE has the lowest PSNR of 14.38 decibels. Overall, MSFT has a significant advantage in restoration accuracy, while other methods make different trade-offs between model complexity, computational efficiency and performance.


\begin{table}[]
\centering
\caption{Params$\downarrow$ and FLOPs$\downarrow$ comparisons with four SOTA methods on the LOL-v1 datasets.}
\fontsize{8}{9}\selectfont
\footnotesize
\begin{tabular}{>{}c>{}c>{}c>{}c>{}c>{}c}
\hline
Methods    & DeepUPE & DeepLPF & SNR-Net & Retinexformer & MSFT  \\ \hline
Params (M) & 1.02    & 1.77    & 4.01    & 1.61          & 1.25  \\
FLOPS (G)  & 21.10   & 5.86    & 26.35   & 15.57         & 18.07 \\
PSNR (dB)  & 14.38   & 15.28   & 24.61   & 22.43         & 27.48 \\ \hline
\label{tab5}   
\end{tabular}
\normalsize
\end{table}

\noindent{\bf{Qualitative Results.}}
The visual comparisons of MSFT and SOTA algorithms on the LOL dataset are shown in Figs. \ref{fig4}, \ref{fig5}, and \ref{fig6}, and enlarge some important detail areas for a better view. In real-world image comparisons, existing methods struggle to capture texture details and tend to introduce dark spots due to enhancement distortion in the illuminated regions of images, as illustrated in the enlarged images of Retinexformer shown in Figure \ref{fig4} and Figure \ref{fig5}, respectively. Our MSFT effectively captures the texture of the background while preserving realistic illumination. In contrast to synthetic images, the enhancement of the extremely dark background also produces a few artifacts, as shown in Figure \ref{fig6}. Our method effectively minimizes artifacts in localized regions.

\begin{table}[t]
    \renewcommand{\arraystretch}{1.25}
    \caption{ Ablation study on the overall structure.}
    \resizebox{1.0\textwidth}{!}
    {
    \begin{tabular}{*{3}{c}*{8}{c}}
    \toprule[0.5mm]
    \multicolumn{3}{c}{Ablation Study}   & \multicolumn{1}{c}{} & \multicolumn{1}{c}{}     & \multicolumn{1}{c}{}  & \multicolumn{1}{c}{}  & \multicolumn{1}{c}{} & \multicolumn{1}{c}{}     & \multicolumn{1}{c}{}   & \multicolumn{1}{c}{}   \\ \cline{1-3}
    \multicolumn{1}{c}{W/O Retinex} & \multicolumn{1}{c}{W/O MSSA} & \multicolumn{1}{c}{W/O FTGT} & \multicolumn{1}{c}{\multirow{-2}{*}{Datasets}} & \multicolumn{1}{c}{\multirow{-2}{*}{LOLv1}} & \multicolumn{1}{c}{\multirow{-2}{*}{LOLv2-real}} & \multicolumn{1}{c}{\multirow{-2}{*}{LOLv2-syn}} & \multicolumn{1}{c}{\multirow{-2}{*}{SID}} & \multicolumn{1}{c}{\multirow{-2}{*}{SMID}} & \multicolumn{1}{c}{\multirow{-2}{*}{SDSD-indoor}} & \multicolumn{1}{c}{\multirow{-2}{*}{SDSD-outdoor}} \\ 
    \midrule[0.5mm]
    &      &      & PSNR& 18.23 & 18.13  & 22.42 & 24.06 & 27.76 & 27.20 & 31.78\\
    \multirow{-2}{*}{\checkmark} & \multirow{-2}{*}{} & \multirow{-2}{*}{}& SSIM & 0.843 & 0.624 & 0.857  & 0.721  &0.826 & 0.869 & 0.898 
    \\ \hline
    &   &   & PSNR & 26.30 & 25.90  & 21.41 & 24.13 & 27.99 & 38.98 & 41.33
    \\
    \multirow{-2}{*}{} & \multirow{-2}{*}{\checkmark}& \multirow{-2}{*}{} & SSIM & 0.854 & 0.818 & 0.745  & 0.720  & 0.819 & 0.792 & 0.987
    \\ \hline
    &   &    & PSNR& 26.98 & 25.87  & 23.47 & 26.03 & 26.81 & 37.73 & 39.10 \\
    \multirow{-2}{*}{}   & \multirow{-2}{*}{} & \multirow{-2}{*}{\checkmark}& SSIM & 0.772 & 0.692 & 0.785  & 0.82  & 0.701 & 0.860 & 0.980 \\ \hline
    &   &    & PSNR& 27.48 & 27.11  & 25.78 & 27.36 & 29.83 & 39.04 & 41.76 \\
    \multirow{-2}{*}{}   & \multirow{-2}{*}{MSFT} & \multirow{-2}{*}{}& SSIM & 0.868 & 0.859 & 0.902  & 0.866  & 0.834 & 0.874 &0.988 \\ 
    \bottomrule[0.5mm]
    \label{Tab2}
    \end{tabular}
    }
\end{table}

\subsection{Ablation Study.}
As shown in Table \ref{Tab2}, we conduct ablation experiments including Retinex ablation, MSSA ablation, and FTGT ablation on the seven datasets. 

\noindent{\bf{Retinex Ablation.}}
We conduct Retinex theory ablation to study the effect of the four-channel illumination map on the performance of MSFT. We remove the fourth channel that contains the brightness information of the normal light map in the initial input. When we apply the Retinex theory, the results of Retinex ablation achieve a large overall improvement of PSNR and SSIM. In particular, on the LOLv1, LOLv2-real, SDSD-indoor, and SDSD-outdoor datasets, improvements in PSNR values are  9.25, 8.98, 11.84, and 9.98dB. Meanwhile, on the LOLv2-real dataset, SSIM achieved the highest performance improvement of 37.7\%, indicating that the application of the Retinex theory in MSFT has strong generalization ability, denoising capability, and illumination restoration in real, noisy complex datasets. The comparative analysis of these data shows the validity of the Retinex theory.

\noindent{\bf{MSSA Ablation.}}
We conduct an ablation study of the MSSA module. The results showed that on the LOLv2-synthetic dataset, removing the MSSA led to a significant drop in PSNR from 25.78 dB to 21.41 dB, and at the same time, SSIM also decreased from 0.902 to 0.745; on the SID dataset, PSNR dropped from 27.36 dB to 24.13 dB, and SSIM also declined from 0.866 to 0.720. The synchronous significant decline in PSNR and SSIM indicates that the MSSA module is not only crucial for enhancing the numerical fidelity of the image, but also has a key impact on restoring the structural similarity and visual naturalness of the image. Its absence will simultaneously damage the objective accuracy and perceptual quality of the image.

\noindent{\bf{Ablation in MSSA.}}
As show in Table \ref{Tab3}, the core of the MSSA block design lies not in the innovation of the unit structure, but in its excellent ability to handle low-resolution sparse information and its lightweight characteristics. This has been verified through the ablation experiments of the system. Firstly, the CMUNeXt unit has significantly improved the PSNR on most datasets. Secondly, the MSA unit performed exceptionally well on the
LOL and SID datasets, with the maximum PSNR and SSIM improvements reaching 1.21 dB and
7.7\% respectively. Finally, the SCSA unit demonstrated strong applicability to the SMID dataset, with a PSNR improvement of 1.65 dB. These data collectively confirm the effectiveness of this design in specific tasks. In addition, as shown in Table \ref{tab7}, we also conducted ablation experiments on the complexity of all sub-modules in MSSA. The experimental data showed that the parameters of MSFT were 1.252MParams, and the computational volume was 18.073GFLOPs. The ablation of SCSA, DSA, or DRA components had almost no impact on the model complexity, and both the parameters and computational volume were almost the same as those of the complete model. In contrast, removing the CMUNeXt component significantly reduced the parameters to 1.101MParams and the computational volume to 17.454GFLOPs. This proves that the sub-modules in MSSA all adopted extremely lightweight designs, and their impact on overall resource consumption was negligible.\\
\noindent{\bf{Ablation in MSA.}}
As show in Table \ref{Tab_DRA_DSA}, MSFT without DRA shows a lower PSNR on LOLv2-real, and the full model achieves an improvement of 1.16 dB, indicating the importance of DRA for this dataset. Thirdly, for DSA, the absence leads to a significant drop in PSNR on SDSD-outdoor, with the full model improving by 3.34 dB, demonstrating the strong applicability of DSA for outdoor scenes.

\begin{table}[!t]
    \renewcommand{\arraystretch}{1.25}

    \caption{Ablation study on the MSSA module}

    \label{Tab3}

    \resizebox{1.0\textwidth}{!}{%

    \begin{tabular}{*{3}{c}*{8}{c}}

    \toprule[0.5mm]

    \multicolumn{3}{c}{Ablation Study} & \multicolumn{1}{c}{} & \multicolumn{1}{c}{} & \multicolumn{1}{c}{} & \multicolumn{1}{c}{} & \multicolumn{1}{c}{} & \multicolumn{1}{c}{} & \multicolumn{1}{c}{} & \multicolumn{1}{c}{} \\ \cline{1-3}

    \multicolumn{1}{c}{W/O SCSA} & \multicolumn{1}{c}{W/O MSA}& \multicolumn{1}{c}{W/O CMUNeXt} & \multicolumn{1}{c}{\multirow{-2}{*}{Datasets}} & \multicolumn{1}{c}{\multirow{-2}{*}{{LOLv1}}} & \multicolumn{1}{c}{\multirow{-2}{*}{LOLv2-real}} & \multicolumn{1}{c}{\multirow{-2}{*}{LOLv2-syn}} & \multicolumn{1}{c}{\multirow{-2}{*}{SID}} & \multicolumn{1}{c}{\multirow{-2}{*}{SMID}} & \multicolumn{1}{c}{\multirow{-2}{*}{SDSD-indoor}} & \multicolumn{1}{c}{\multirow{-2}{*}{SDSD-outdoor}} \\ 

    \midrule[0.5mm]

     & & & PSNR & 27.41 & 26.27 & 25.63 & 26.59 & 28.18 & 38.86 & 41.60 \\

 \multirow{-2}{*}{\checkmark} & \multirow{-2}{*}{} & \multirow{-2}{*}{} & SSIM & 0.830 & 0.850 & 0.793 & 0.862 & 0.820 & 0.870 & 0.988 \\ \hline

     & & & PSNR & 26.94 & 25.95 & 25.55 & 26.54 & 29.80 & 39.04 & 41.75\\

 \multirow{-2}{*}{} & \multirow{-2}{*}{\checkmark} & \multirow{-2}{*}{} & SSIM & 0.837 & 0.789 & 0.810 & 0.874 & 0.830 & 0.873 & 0.988 \\ \hline

     & & & PSNR & 27.39 & 26.41 & 25.26 & 25.99 & 29.01 & 38.20 & 38.55\\

    \multirow{-2}{*}{} & \multirow{-2}{*}{} & \multirow{-2}{*}{\checkmark} & SSIM & 0.811 & 0.849 & 0.888 & 0.723 & 0.800 & 0.841 & 0.976 \\ \hline

     & & & PSNR & 27.48 & 27.11 & 25.78 & 27.36 & 29.83 & 39.04 & 41.76 \\

    \multicolumn{3}{c}{\multirow{-2}{*}{MSFT}} & SSIM & 0.868 & 0.859 & 0.902 & 0.866 & 0.834 & 0.874 & 0.988 \\ 

    \bottomrule[0.5mm]

    \end{tabular}

    }

\end{table}

\begin{table}[!t]
    \renewcommand{\arraystretch}{1.25}
    \caption{Ablation study on the DRA and DSA modules}
    \label{Tab_DRA_DSA}
    \resizebox{1.0\textwidth}{!}{%
    \begin{tabular}{*{2}{c}*{8}{c}}
    \toprule[0.5mm]
    \multicolumn{2}{c}{Ablation Study} & \multicolumn{1}{c}{} & \multicolumn{1}{c}{} & \multicolumn{1}{c}{} & \multicolumn{1}{c}{} & \multicolumn{1}{c}{} & \multicolumn{1}{c}{} & \multicolumn{1}{c}{} & \multicolumn{1}{c}{} \\ \cline{1-2}
    \multicolumn{1}{c}{W/O DRA} & \multicolumn{1}{c}{W/O DSA} & \multicolumn{1}{c}{\multirow{-2}{*}{Datasets}} & \multicolumn{1}{c}{\multirow{-2}{*}{LOLv1}} & \multicolumn{1}{c}{\multirow{-2}{*}{LOLv2-real}} & \multicolumn{1}{c}{\multirow{-2}{*}{LOLv2-syn}} & \multicolumn{1}{c}{\multirow{-2}{*}{SID}} & \multicolumn{1}{c}{\multirow{-2}{*}{SMID}} & \multicolumn{1}{c}{\multirow{-2}{*}{SDSD-indoor}} & \multicolumn{1}{c}{\multirow{-2}{*}{SDSD-outdoor}} \\ 
    \midrule[0.5mm]

     & & PSNR & 26.94 & 25.95 & 25.55 & 26.54 & 29.80 & 39.04 & 41.75 \\
    \multirow{-2}{*}{\checkmark} & \multirow{-2}{*}{} & SSIM & 0.837 & 0.789 & 0.810 & 0.864 & 0.830 & 0.873 & 0.988 \\ \hline

     & & PSNR & 27.32 & 25.98 & 25.54 & 26.40 & 28.94 & 37.94 & 38.42\\
    \multirow{-2}{*}{} & \multirow{-2}{*}{\checkmark} & SSIM & 0.821 & 0.841 & 0.891 & 0.724 & 0.810 & 0.838 & 0.977 \\ \hline

     & & PSNR & 27.48 & 27.11 & 25.78 & 27.36 & 29.83 & 39.04 & 41.76 \\
    \multicolumn{2}{c}{\multirow{-2}{*}{MSFT}} & SSIM & 0.868 & 0.859 & 0.902 & 0.866 & 0.834 & 0.874 & 0.988 \\ 

    \bottomrule[0.5mm]
    \end{tabular}
    }
\end{table}

\noindent{\bf{FTGT Ablation.}}
In the ablation experiment of the FTGT module, removing this module led to a general decline in the performance of all datasets. Among them, the reduction in PSNR and SSIM was particularly significant. For instance, on the LOLv2-real dataset, PSNR decreased from 27.11 dB to 25.87 dB, and SSIM dropped from 0.859 to 0.692, with a relative decrease of approximately 24.13\%; on the LOLv2-synthetic dataset, PSNR decreased from 25.78 dB to 23.47 dB, and SSIM dropped from 0.902 to 0.785, with a relative decrease of about 14.90\%; on the SMID dataset, PSNR decreased from 29.83 dB to 26.81 dB, and SSIM dropped from 0.834 to 0.701, with a relative decrease of approximately 18.97\%. Other datasets such as LOLv1 and SID also showed similar trends, with relative decrease percentages of 12.44\% and 5.61\% respectively for SSIM.

\noindent{\bf{Ablation in FTGT.}}
As shown in Table \ref{tabFTGT}, we conducted ablation experiments on the core modules FFT and FFN in the FTGT. The experimental results show that both modules are crucial for performance improvement, but the FFT module plays a more core role. When the FFT module is ablated, the performance of MSFT drops sharply on all datasets, especially on the SDSD-outdoor dataset, where PSNR drops from 41.76 to 34.98, and SSIM drops from 0.988 to 0.880. This proves that the FFT module is indispensable for capturing and reconstructing the global frequency information and structural consistency of the image. In contrast, ablation of the FFN module also leads to performance degradation, but to a lesser extent. For example, on the LOLv1 dataset, PSNR drops from 27.48 to 25.97, and SSIM drops from 0.868 to 0.798. This indicates that FFN focuses more on the refinement of local features and non-linear transformation. Overall, FFT is the foundation for the model to achieve high fidelity results, while FFN optimizes and supplements on this basis.

\begin{table}[!t]
    \renewcommand{\arraystretch}{1.25}
    \caption{Ablation study on the FTGT module}
    \resizebox{1.0\textwidth}{!}
    {
    \begin{tabular}{*{2}{c}*{8}{c}}
    \toprule[0.5mm]
    \multicolumn{2}{c}{Ablation Study} & \multicolumn{1}{c}{} & \multicolumn{1}{c}{} & \multicolumn{1}{c}{} & \multicolumn{1}{c}{} & \multicolumn{1}{c}{} & \multicolumn{1}{c}{} & \multicolumn{1}{c}{} & \multicolumn{1}{c}{} \\ \cline{1-2}
    \multicolumn{1}{c}{W/O FFT} & \multicolumn{1}{c}{W/O FFN} & \multicolumn{1}{c}{\multirow{-2}{*}{Datasets}} & \multicolumn{1}{c}{\multirow{-2}{*}{LOLv1}} & \multicolumn{1}{c}{\multirow{-2}{*}{LOLv2-real}} & \multicolumn{1}{c}{\multirow{-2}{*}{LOLv2-syn}} & \multicolumn{1}{c}{\multirow{-2}{*}{SID}} & \multicolumn{1}{c}{\multirow{-2}{*}{SMID}} & \multicolumn{1}{c}{\multirow{-2}{*}{SDSD-indoor}} & \multicolumn{1}{c}{\multirow{-2}{*}{SDSD-outdoor}} \\ 
    \midrule[0.5mm]

    & & PSNR & 24.41 & 23.83 & 23.70 & 22.46 & 24.54 & 35.59 & 34.98\\
    \multirow{-2}{*}{\checkmark} & \multirow{-2}{*}{} & SSIM & 0.701 & 0.697 & 0.732 & 0.654 & 0.674 & 0.768 & 0.880 \\ \hline

    & & PSNR & 25.97 & 24.98 & 24.78 & 25.94 & 27.84 & 36.45 & 37.17\\
    \multirow{-2}{*}{} & \multirow{-2}{*}{\checkmark} & SSIM & 0.798 & 0.801 & 0.756 & 0.691 & 0.724 & 0.803 & 0.879 \\ \hline

    & & PSNR & 27.48 & 27.11 & 25.78 & 27.36 & 29.83 & 39.04 & 41.76 \\
    \multicolumn{2}{c}{\multirow{-2}{*}{MSFT}} & SSIM & 0.868 & 0.859 & 0.902 & 0.866 & 0.834 & 0.874 & 0.988 \\ 

    \bottomrule[0.5mm]
    \label{tabFTGT}
    \end{tabular}
    }
\end{table}

\newcolumntype{P}{c}

\begin{table}[]
\centering
\caption{Ablation Study on the Complexity of the MSSA Module}
\begin{tabular}{PPPPPP}
\hline
Methods    & w/o SCSA & w/o DSA & w/o DRA & w/o CMUNeXt & MSFT  \\ \hline
Params (M) & 1.250     & 1.242    & 1.246    & 1.101        & 1.252  \\
FLOPS (G)  & 18.073   & 18.072    & 18.071   & 17.454       & 18.073 \\ \hline
\label{tab7}
\end{tabular}
\end{table}

\section{Conclusion}
We propose a Transformer-based method combined with the Fourier transform for low-light image enhancement, which conducts two-stage guidance following the extraction of attention weight features and the fusion of frequency domain information within a dual convolutional channel. Our approach is capable of effectively enhancing the background details of low-light images while accurately restoring the original illumination. In addition, our method provides a reference for the construction of models that use a frequency-domain guided attention mechanism to restore degraded images. Our framework integrates multi-scale feature extraction, enhancing its effectiveness for image recovery and model performance.

\noindent \textbf{Limitations}
Although our method achieves good results in enhancing image quality in extremely dark environments, it is limited in dealing with images polluted by strong light due to the ineffective removal of light pollution noise from the exposed part, and is not sufficiently efficient for real-time processing of natural light. This trained model requires a large amount of paired data for training to perform well. Therefore, it is not suitable for an unsupervised augmentation field without paired data at present. In addition, this model does not take into account images acquired in scenarios with severe noise pollution. It requires a relatively clean paired dataset or necessitates prior cleansing of the dataset before training.

\noindent \textbf{Future work}
This research specifically focuses on the cross-attention mechanism and the CNN-Transformer hybrid architecture in frequency-domain learning. However, it fails to truly integrate the Fourier transform into the attention layer and apply attention in the frequency domain. For instance, performing Attention calculations using the amplitude spectrum and phase spectrum can further explore the superior performance of the Transformer in frequency-domain learning. Secondly, the application of the diffusion model in the frequency domain has not been fully explored. Although its powerful generative ability performs excellently in the field of unsupervised low-light enhancement, it fails to fully demonstrate its capabilities due to the high computational resources required by its complex model. It is hoped that future work can further explore its superior performance by optimizing the structure of the diffusion model and reducing computational resources. Additionally, the superior performance of the Transformer in frequency-domain learning can be further explored.

\section*{Acknowledgements}
This work was partially supported by the Natural Science Foundation of Shanghai under Grant 25ZR1402098, the Shanghai Science and Technology project under Grant 24LZ1401600, the National Natural Science Foundation of China under Grant 62373155, and the startup fund from East China University of Science and Technology under Grant YH0142234.

\bibliographystyle{elsarticle-num.bst}
\bibliography{reference.bib}

\end{document}